# *HOG Based fast Human Detection*

M. Kachouane, S. Sahki
Electronics dept.
Université Saad Dahlab de Blida
Blida – Algeria
Kachouane.mouloud@hotmail.fr

M. Lakrouf, N. Ouadah
NCRM team, Robotics & Automation dept.
Centre de Développement des Technologies Avancées
Baba Hassen, Algiers, Algeria
mlakrouf@cdta.dz

*Abstract*—Objects recognition in image is one of the most difficult problems in computer vision. It is also an important step for the implementation of several existing applications that require high-level image interpretation. Therefore, there is a growing interest in this research area during the last years. In this paper, we present an algorithm for human detection and recognition in real-time, from images taken by a CCD camera mounted on a car-like mobile robot. The proposed technique is based on Histograms of Oriented Gradient (HOG) and SVM classifier. The implementation of our detector has provided good results, and can be used in robotics tasks.

*Human detection; HOG descriptor; SVM classifier; Real- time detection.*

I. INTRODUCTION

The problem of object recognition is to decide whether a specific object or object within a class of objects is contained in an image or not. This problem can be seen as a match between the target model and a set of descriptors, which are extracted from an image test. This generalization, as simple as it seems to be, is able to explain the existence of multiple approaches that depends on the choice of object descriptors, type and complexity of its model, and the methods used for the learning and matching object model. Beyond being a general problem in computer vision, object recognition is an important tool for many applications. It is used in video surveillance, digital image databases, and largely for complex robotic tasks as in our case.

Person detection is particularly difficult, mainly because of the high variability of appearances and possible situations. The problem is to find a representation of a human that is both sufficiently generic to cover all types of situations, and sufficiently discriminative for humans. For this, we generally use an intermediate representation, based on the computation of one or more features, taken from the information contained in the only values of the image pixels.

In this paper, we address the problem of person detection from images taken by a CCD camera, embedded on an outdoor mobile robot. Several approaches have been implemented in real-time, to perform an efficient and fast detection, in order to be used in autonomous navigation tasks. This work has been done in the context of autonomous transportation system project of NCRM team, in CDTA.

The paper is organized as follow: in section II we present the HOG descriptor combined with the SVM classifier, used in this work. Section III is dedicated to show experimental tests and results discussion.

II. HUMAN DETECTION APPROCH

Early works on people detection date from the late 1990s [1]. In one of the first proposed methods, stereovision is used to detect objects using a Hough transform. The method can detect pedestrians, but is not exclusive to this type of object. In 1998, Wöhler and Heisele use leg movement to achieve the pedestrian detection and classification, with constraints on the location of pedestrians to the ground [1]. However, these methods are not generic.

From the 2000s, research advances focused more on face detection and in particular the method of Viola & Jones, which has been extended in 2005 to target detection using motion [2]. The method allows a more generic detection, requiring no prior information about the structure of the scene, with a processing time close to the real-time. In 2005, INRIA researchers propose a new technique based on the histograms of oriented gradient (HOG) [3]. The good performances obtained by this technique make it few years after a standard method [4]. In 2008, researchers from Rutgers University have introduced a new descriptor built as a covariance matrix, which provides better performance in known environment [5]. Also in 2008, Mu & al. have successfully used local binary patterns (LBP), a type of features that had proven a good effectiveness especially when applied to face detection [6]. In this work, we focus our interest on the HOG based technique, which is detailed bellow:

*A. HOG descriptor*

HOG descriptors have been introduced by Dalal and Triggs in [7] and [8]. The main idea behind the histogram of oriented gradient is that the local appearance and shape of object in an image can be described by the intensity distribution of gradients or direction of the contours. The implementation of these descriptors can be obtained by dividing the image into small connected regions, called cells. Then, for each cell we compute a histogram of gradient directions or edge orientations for all pixels of the cell. The combination of these histograms is the descriptor.

The HOG descriptor has some key advantages. Since it operates on localized cells, the method maintains the invariance to geometric and photometric transformations. We implemented the HOG descriptor on C/C++ and using OpenCV [9] libraries following four steps as shown below:

*1) Training database conception:* the database used was the "INRIA pedestrian database" which contains images covering a wide variety of pedestrians. We have then updated it by adding images of pedestrians in different states (running, standstill, in front, side and back) and some images of pedestrians in clothes from Islamic society: women wearing veil of different colors and shapes.

*2) Luminance normalisation:* The input images of our system are initially converted to grayscale. Then their luminance is normalized. We used then the Gamma correction.

*3) Image gradient :* The gradient is a key step for the descriptors formation. The accuracy of computed orientations and histograms and the results are closely related to the method used to calculate the gradient of the image. Fast computation of the gradient can be done using, for example, 1-D simple derivation masks, 2-D operators like Sobel or recursive operators like Deriche.

In our case, we have used the algorithm of the first derivative which is one of the most simple and fast operators. This operator uses convolution matrix to calculate an approximation of the horizontal and vertical derivative. Let I be the source image. Images which contain at each point the derivative approximations respectively horizontal and vertical are calculated as follows:

$$G_x = [-1\ 0\ 1] \times I \quad (1)$$

$$G_y = [-1\ 0\ 1]' \times I \quad (2)$$

At each point, the approximations of horizontal and vertical gradients are combined as follows to obtain an approximation of the gradient norm:

$$G = \sqrt{G_x^2 + G_y^2} \quad (3)$$

We also compute the gradient direction as follows:

$$\theta = arctan\left(\frac{G_y}{G_x}\right) \quad (4)$$

*4) Gradient orientation histograms*: A histogram is an array of numbers where each element corresponds to the frequency of occurrence of a range of values for a set of data. In a part of an image, for example, each box of the histogram may represent the pixels with the same color. A histogram is a transformation of the data space to the positive real numbers. From a statistical point of view, a histogram provides distribution of a certain type of data set. The image is divided into cells of 8x8 pixels size, and for each cell we compute the gradient orientation histogram. Each pixel of the cells participates in the vote. For each pixel of coordinates (x, y), the associated value in the histogram H is given by:

$$H(a) = H(a) + NG(x, y) \quad (5)$$

We have created 9 sub-images called binaries images, as we have chosen the size of the histogram box equal to 20° (180°/20° = 9). All pixels in these images are set to zero except the pixels in the original image for which the values of the gradient orientation correspond to the particular case. These 9 images constitute the full histogram. For each cell of size 8 x 8 pixels, we compute the values of the 9 boxes of gradient orientation histogram. For a block of 2 x 2 cells, we calculate the HOG descriptor for each cell, then the obtained arrays are assembled into a single array of 36 components, and it is normalized in accordance with standard L1 defined by the normalization factor f given by:

$$f = \frac{v}{(\|v\|_1 + \epsilon)} \quad (6)$$

were v represents the non-normalized vector containing all the histograms of a single block, $\|v\|_k$ its k-norm and ε is a constant.

This normalized vector corresponds to the HOG descriptor for a block. An image 64x128 pixels, contains 7 x 15 blocks with multiple overlap. We assemble the normalized vectors obtained for all blocks in a single 1-D vector of 3780 components, which represents the final HOG descriptor for our case.

*B. SVM Classifier*

The set of descriptors (105 x 36 = 3780 values) is used to feed the SVM classifier, which generates a model (a set of support arrays). During the decision phase, the descriptors are calculated in an identical manner as in the learning phase. Decision making, regarding the class membership is made directly by the decision function of SVM. In our experiments, we use the linear kernel function to classify the descriptors given by the learning database.

III. EXPERIMENTAL RESULTS

We have implemented the HOG based detection technique in real-time on the mobile robot "Robucar" available in CDTA, under two different environments: a wide corridor (indoor environment) and a parking (outdoor environment). For that, we have used our database (called CDTA base), composed by 112 images of 320x240 pixels, and containing 216 human targets. In this case, we have considered as being a good detection if the frame covers more than the half of the human body. Images acquisitions and processing management are performed within a modular LAAS-CNRS architecture called "GenoM" [10], using C/C++ and the library OpenCV on a laptop with 2 GHz Core-i5 processor and 4 GB Ram.

Detection parameters of the proposed technique have been tested: Gamma correction of the image, gradient filter type, blocks and cells sizes and the global threshold. The goal is to identify the best parameters to increase the detection rate, and decrease the processing time.

## A. Image Gamma correction

TABLE I. DETECTION RATE & GAMMA CORRECTION

|  | Detection | False positive | Processing time (ms) |
|---|---|---|---|
| With Correction | 125 (58%) | 17 (8%) | 200 |
| Without Correction | 105 (49%) | 11 (5%) | 180 |

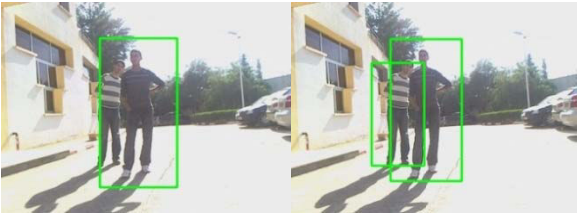

Figure 1. Human detection without (left) and with (right) Gamma correction

We notice that introducing the Gamma correction of the image enhances the detection rate.

## B. Gradient filter

We have tested two gradient filters, and here are the results:

TABLE II. DETECTION RATE & GRADIENT'S FILTERS

| Gradient filter | Detection | False positive | Processing time (ms) |
|---|---|---|---|
| Sobel | 127 (59%) | 16 (7.5%) | 240 |
| 1-D Derivator | 125 (58%) | 17 (8%) | 180 |

Note that the Sobel filter gives slightly better results, but with less processing time. We chose then to use the 1-D differentiator, because it is faster.

## C. Cells & Blocks sizes :

TABLE III. DETECTION RATE & CELLS SIZE

| Cells size (pixel) | Detection | False positive | Processing time (ms) |
|---|---|---|---|
| 6 x 6 | 162 (75%) | 20 (9%) | 700 |
| 7 x 7 | 163 (75%) | 17 (8%) | 600 |
| 8 x 8 | 158 (73%) | 17 (8%) | 150 |
| 9 x 9 | 158 (73%) | 16 (7%) | 1000 |
| 10 x 10 | 138 (64%) | 13 (6%) | 547 |

The cells size selected affects the quality of detection, and even more on the detection time. We chose the 8x8 size for a reasonable amount of time. The sizes below are slightly better, but their computation time does not allow a real-time application.

TABLE IV. DETECTION RATE & BLOCKS SIZE

| Block size | Detection | False positive | Processing time (ms) |
|---|---|---|---|
| 24 x 24 | 160 (74%) | 16 (7%) | 135 |
| 28 x 28 | 168 (78%) | 17 (8%) | 143 |
| 30 x 30 | 172 (80%) | 16 (7%) | 145 |
| 32 x 32 | 173 (80%) | 17 (8%) | 150 |
| 36 x 36 | 177 (82%) | 18 (8%) | 160 |

We have then tried several combinations of blocks size and cells size. The one which gives the best detection is: blocks of 36x36 pixels, 9x9 pixels cell, but it constrains execution time to be near to 3 seconds. For real-time use (as for robotics tasks), we have chosen the combination: 32x32, 8x8, as it gives a quite good detection, with low processing time (135ms ≈ 7 frames/second).

## D. Global threshold

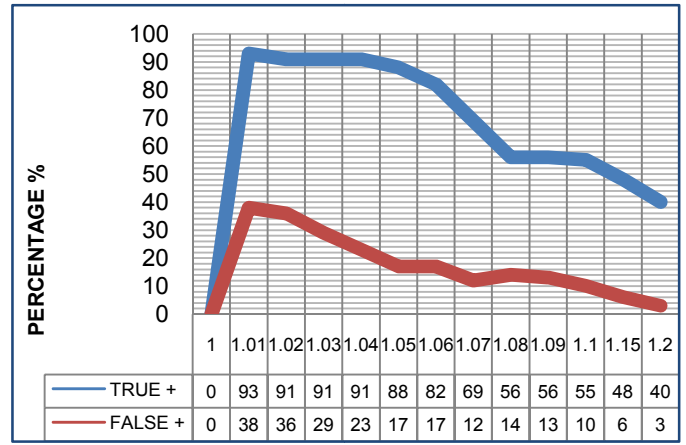

Figure 2. Detection rate according to global threshold

From this figure, we notice that the best detection rates (in bleu) are between thresholds values of 1.01 and 1.06, and the rate of false detection is quite low (in yellow) from 1.04.

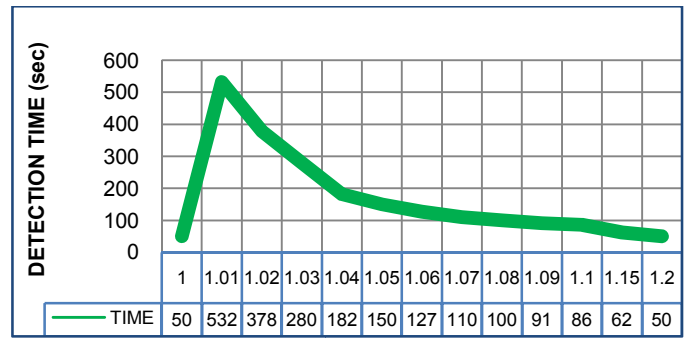

Figure 3. Detection time according to global threshold

In the second figure, we can see that processing time takes lowest values when the threshold is high. Combining both results, we have chosen the value of 1.05 which gives a fairly high detection rate. The false positive rate is low where detection time is close to be on real-time.

After parameters evaluations and tunings, we have tested our programs on several databases. Among them, one downloaded from the Internet (INRIA database). Here are presented some detection results applied on images taken from this database.

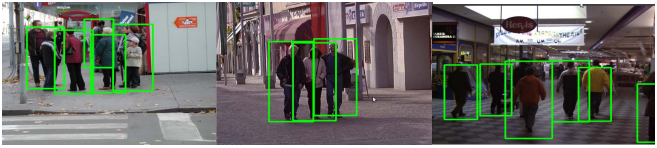

Figure 4. HOG based detector on INRIA database

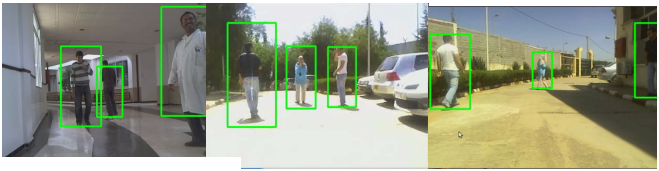

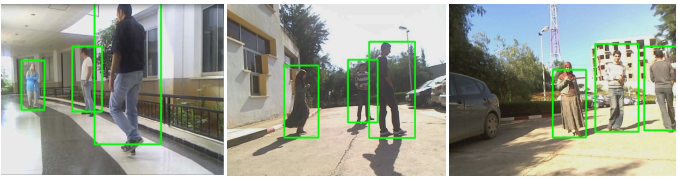

Figure 5. HOG based detector on CDTA database

Figure 5 presents detection results applied on images taken from our base. Are summarized in this table statistics on detection rates and false detections taken from tests on the two cited databases.

| Databases | Image number | Target number | Detected target | False detection | Rate detection |
|---|---|---|---|---|---|
| **INRIA** | 900 | 1178 | 1012 | 170 (17%) | 86% |
| **CDTA** | 1689 | 4031 | 3491 | 712 (17%) | 87% |
| **All** | 2589 | 5209 | 4503 | 822 (17%) | 86% |

TABLE V. TABLE OF RESULTS

## IV. CONCLUSION

In this paper, we present fast humans detection system implemented, with our own database, using histograms of oriented gradient (HOG) combined with the SVM classifier. Several tests have been performed to identify optimal parameters. Global detection rate is quite interesting, form results of detector application on two different databases. Processing time is suitable for embedded applications, but can be enhanced by detector combination. Future works will focus on integrating this technique in complex robotic task.

ACKNOWLEDGMENT

The authors are very grateful to Anthony Mallet from LAAS-CNRS, for its involvements in this work.